\documentclass[letterpaper]{article} 
\usepackage{aaai24}  
\usepackage{times}  
\usepackage{helvet}  
\usepackage{courier}  
\usepackage[hyphens]{url}  
\usepackage{graphicx} 
\usepackage{natbib}  
\usepackage{caption} 
\usepackage{algorithm}
\usepackage{algorithmic}

\usepackage{multicol}
\usepackage{multirow}
\usepackage{amssymb}
\usepackage{amsmath}
\usepackage{newfloat}
\usepackage{listings}
\usepackage{bbding}
\usepackage{pifont}
\usepackage{stfloats}
\DeclareMathOperator*{\argmax}{arg\,max}
\DeclareMathOperator*{\argmin}{arg\,min}

\usepackage{newfloat}
\usepackage{listings}
\DeclareCaptionStyle{ruled}{labelfont=normalfont,labelsep=colon,strut=off} 
\floatstyle{ruled}
\newfloat{listing}{tb}{lst}{}
\floatname{listing}{Listing}
\title{Unsupervised Continual Anomaly Detection with Contrastively-learned Prompt}
\author {
    Jiaqi Liu\textsuperscript{\rm 1}\thanks{Contributed Equally.},
    Kai Wu\textsuperscript{\rm 2}$^*$,
    Qiang Nie\textsuperscript{\rm 2},
    Ying Chen\textsuperscript{\rm 2},
    Bin-Bin Gao\textsuperscript{\rm 2},\\
    Yong Liu\textsuperscript{\rm 2},
    Jinbao Wang\textsuperscript{\rm 1},
    Chengjie Wang\textsuperscript{\rm 2,3},
    Feng Zheng\textsuperscript{\rm 1}\thanks{Corresponding Author.}
}
\affiliations {
    \textsuperscript{\rm 1}Southern University of Science and Technology\\
    \textsuperscript{\rm 2}Tencent Youtu Lab\\
    \textsuperscript{\rm 3}Shanghai Jiao Tong University\\
    liujq32021@mail.sustech.edu.cn, lloydwu@tencent.com, stephennie@tencent.com, mumuychen@tencent.com,\\ 
    csgaobb@gmail.com,
    choasliu@tencent.com, linkingring@163.com, jasoncjwang@tencent.com, zfeng02@gmail.com
}

\usepackage{bibentry}

\newcommand{\squishlist}{
 \begin{list}{$\bullet$}
  { \setlength{\itemsep}{0pt}
     \setlength{\parsep}{1pt}
     \setlength{\topsep}{1pt}
     \setlength{\partopsep}{0pt}
     \setlength{\leftmargin}{1.5em}
     \setlength{\labelwidth}{1em}
     \setlength{\labelsep}{0.5em} } }
\newcommand{\squishend}{
  \end{list}  }

\begin{document}

\maketitle

\begin{abstract}
Unsupervised Anomaly Detection (UAD) with incremental training is crucial in industrial manufacturing, as unpredictable defects make obtaining sufficient labeled data infeasible. 
However, continual learning methods primarily rely on supervised annotations, while the application in UAD is limited due to the absence of supervision.
Current UAD methods train separate models for different classes sequentially, leading to catastrophic forgetting and a heavy computational burden.
To address this issue, we introduce a novel Unsupervised Continual Anomaly Detection framework called \textbf{UCAD}, which equips the UAD with continual learning capability through contrastively-learned prompts. 
In the proposed UCAD, we design a Continual Prompting Module (CPM) by utilizing a concise key-prompt-knowledge memory bank to guide task-invariant `anomaly' model predictions using task-specific `normal' knowledge.
Moreover, Structure-based Contrastive Learning (SCL) is designed with the Segment Anything Model (SAM) to improve prompt learning and anomaly segmentation results.
Specifically, by treating SAM's masks as structure, we draw features within the same mask closer and push others apart for general feature representations.
We conduct comprehensive experiments and set the benchmark on unsupervised continual anomaly detection and segmentation, demonstrating that our method is significantly better than anomaly detection methods, even with rehearsal training. 
The code will be available at \url{https://github.com/shirowalker/UCAD}.

%

\end{abstract}

\begin{figure}[t]
\centering
\includegraphics[width=1.0\columnwidth]{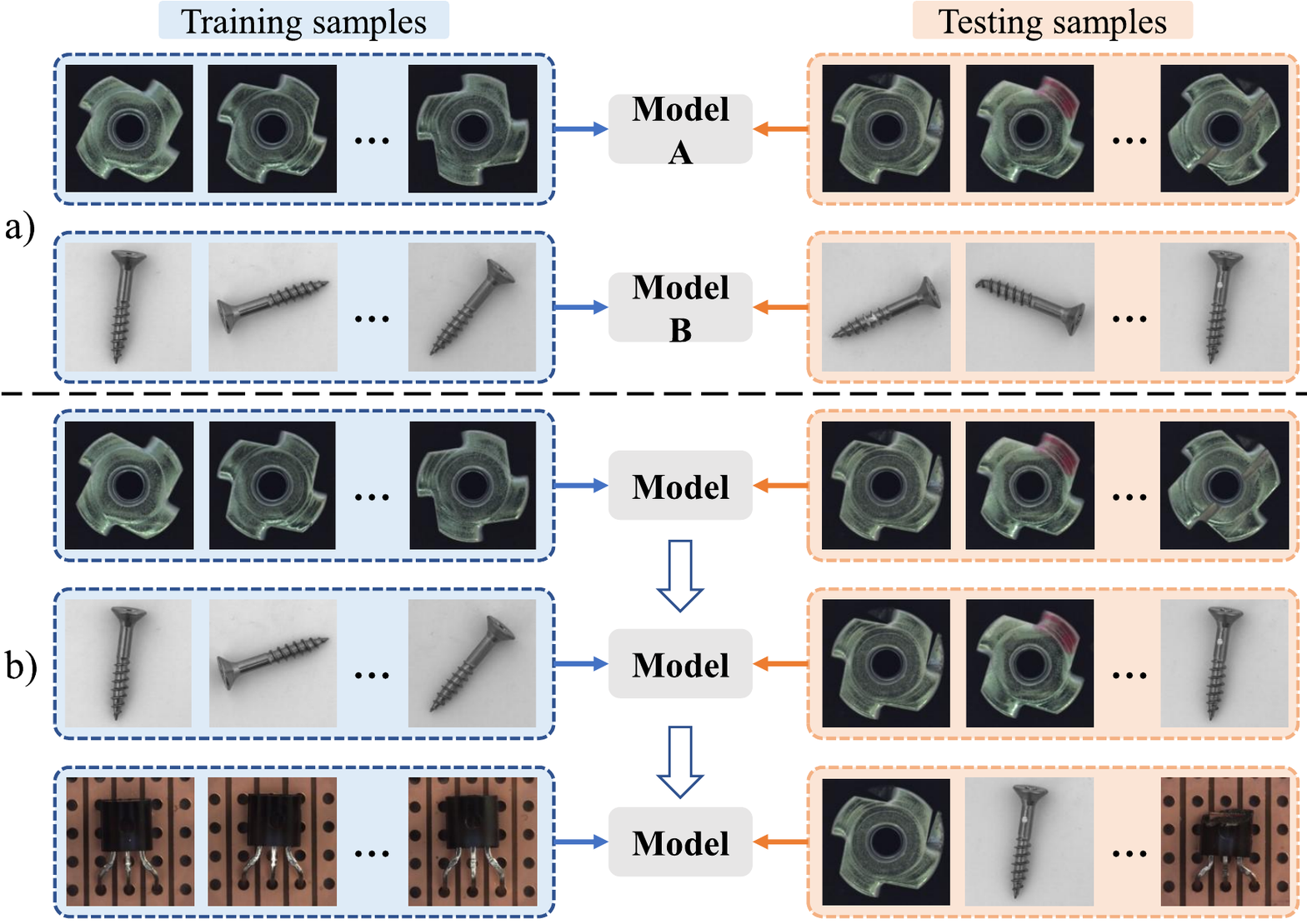} 
\caption{Comparison between separate models and UCAD methods: a) Using separate methods, each task has its own individual model. On the contrary, Ours b) uses a single model to handle all tasks without task identities. In the continuous stream, UCAD only requires the dataset of the current task for training and can be applied to previous tasks.}
\label{fig1}
\end{figure}

\section{Introduction}


Unsupervised Anomaly Detection (UAD) focuses on identifying unusual patterns or outliers in data without prior knowledge or labeled instances, relying solely on the inherent distribution of the `normal' data \cite{chandola2009anomaly}. 
This approach is particularly useful in industrial manufacturing since acquiring well-labeled defect data can be challenging and costly. 

Recent researches on UAD involve training distinct models for various classes, which inevitably relies on the knowledge of class identity during the test phase ~\cite{liu2023deep}. 
Moreover, forcing separate models to learn sequentially also results in a heavy computational burden with class incrementation.
Some other methods focus on training a unified model that can handle multiple classes, such as UniAD~\cite{you2022unified}. 
In real production, trains occur sequentially, which makes it impractical for UniAD to require all data to be trained simultaneously.
Additionally, the unified model still lacks the ability to retain previously learned knowledge when continuously adapting to 
frequent product alterations during sequential training.
Catastrophic forgetting and computational burden hinder UAD methods from applying to real-world scenarios. 

Continual Learning (CL) is well-known for addressing the issue of catastrophic forgetting with a single model, especially when previous data is unavailable due to privacy reasons~\cite{li2023cross}.
Recent research on continual learning can be categorized based on the requirement of task identities during the test phase.
Task-aware approaches explicitly use the task identities to guide the learning process and prevent interference between tasks \cite{aljundi2018memory, kirkpatrick2017overcoming}.
However, it's not always possible to acquire task identities during inference. Hence, task-agnostic methods are necessary and more prevail. 
 \citeauthor{aljundi2019task} progressively modifies data distribution to adapt various tasks in an online setup.
L2P \cite{wang2022learning} dynamically learns prompts as task identities. 
Despite the effectiveness of task-agnostic CL methods in supervised tasks, their efficacy in UAD remains unproven. 
Obtaining large scales of anomalous data is difficult in industries due to high production success rates and privacy concerns. 
Therefore, it is crucial to explore the application of CL in UAD.

To date, there is no known effort, except for Gaussian distribution estimator (DNE) \cite{li2022towards}, to incorporate CL into UAD.
However, DNE still relies on augmentations~\cite{li2021cutpaste} to provide pseudo-supervision and is not applicable to anomaly segmentation. 
DNE can be considered a continual binary image classification method rather continual anomaly detection (AD) method. 
In real industrial manufacturing, accurately segmenting the areas of anomalies is essential for anomaly standard quantization. 
Hence, there is an urgent need for a method that can perform unsupervised continual AD and segmentation simultaneously.

To address the aforementioned problems, we propose a novel framework for Unsupervised Continual Anomaly Detection called UCAD, which can sequentially learn to detect anomalies of different classes using a single model, as shown in Fig.~\ref{fig1}.
UCAD incorporates a Continual Prompting Module (CPM) to enable CL in unsupervised AD and a Structure-based Contrastive Learning (SCL) module to extract more compact features across various tasks. The CPM learns a ``key-prompt-knowledge" memory space to store auto-selected task queries, task adaptation prompts, and the `normal' knowledge of different classes. Given an image, the key is automatically selected to retrieve the corresponding task prompts. Based on the prompts, the image feature is further extracted and compared with its normal knowledge for anomaly detection, similar to PatchCore~\cite{roth2022towards}.
However, the performance of CPM is limited because the frozen backbone (ViT) cannot provide compact feature representations across various tasks. 
To overcome this limitation, the SCL is introduced to extract more dominant feature representations and reduce domain gaps by leveraging the general segmentation ability of SAM~\cite{Kirillov_2023_ICCV}. 
With SCL, features of the same structure (segmented area) are pulled together and pushed away from features in other structures. As a result, the prompts are contrastively learned for better feature extraction across different tasks.
Our contributions can be summarized as follows:
\squishlist

\item To the best of our knowledge, our proposed UCAD is the first framework for task-agnostic continual learning on unsupervised anomaly detection and segmentation. UCAD novelty learns a key-prompt-knowledge memory space for automatic task instruction, knowledge transfer, unsupervised anomaly detection and segmentation.  
\item We propose to use contrastively-learned prompts to improve unsupervised feature extraction among various classes by exploiting the general capabilities of SAM. 
\item We have conducted thorough experiments and introduced a new benchmark for unsupervised CL anomaly detection and segmentation. Our proposed UCAD outperforms previous state-of-the-art (SOTA) AD methods by 15.6\% on detection and 26.6\% on segmentation. 
\squishend

\section{Related Work}
 \subsection{Unsupervised Image Anomaly Detection}
With the release of the MVTec AD dataset~\cite{bergmann2019mvtec}, the development of industrial image anomaly detection has shifted from a supervised paradigm to an unsupervised paradigm. In the unsupervised anomaly detection paradigm, the training set only consists of normal images, while the test set contains both normal images and annotated abnormal images. Gradually, research on unsupervised industrial image anomaly detection has been divided into two main categories: feature-embedding-based methods and reconstruction-based methods~\cite{liu2023deep}. \textbf{Feature-embedding-based methods} can be further categorized into four subcategories, including \textit{teacher-student model}~\cite{bergmann2020uninformed,salehi2021multiresolution,deng2022anomaly,tien2023revisiting}, \textit{one-class classification methods}~\cite{li2021cutpaste,liu2023simplenet}, \textit{mapping-based methods}~\cite{rudolph2021same,gudovskiy2022cflow,rudolph2022fully,lei2023pyramidflow} and \textit{memory-based methods}~\cite{defard2021padim,roth2022towards,jiang2022softpatch,xie2022pushing,liu2023real3d}. \textbf{Reconstruction-based methods} can be categorized based on the type of reconstruction network, including \textit{autoencoder-based methods}~\cite{zavrtanik2021draem,zavrtanik2022dsr,schluter2022natural}, \textit{Generative Adversarial Networks (GANs)~\cite{goodfellow2014generative} based methods}~\cite{yan2021learning,liang2022omni}, \textit{ViT-based methods}~\cite{mishra2021vt,pirnay2022inpainting,jiang2022masked}, and \textit{Diffusion model-based methods}~\cite{mousakhan2023anomaly,zhang2023diffusionad}.

However, existing UAD methods are designed to enhance anomaly detection capabilities within a single object category. They often lack the ability to perform anomaly detection in a continual learning scenario. Even multi-class unified anomaly detection models~\cite{you2022unified,zhao2023omnial} have not taken into consideration the scenario of continual anomaly detection. While our method is specifically designed for the scenario of continual learning and achieves continual anomaly segmentation in an unsupervised manner.

\begin{figure*}[t]
\centering
\includegraphics[width=2.0\columnwidth]{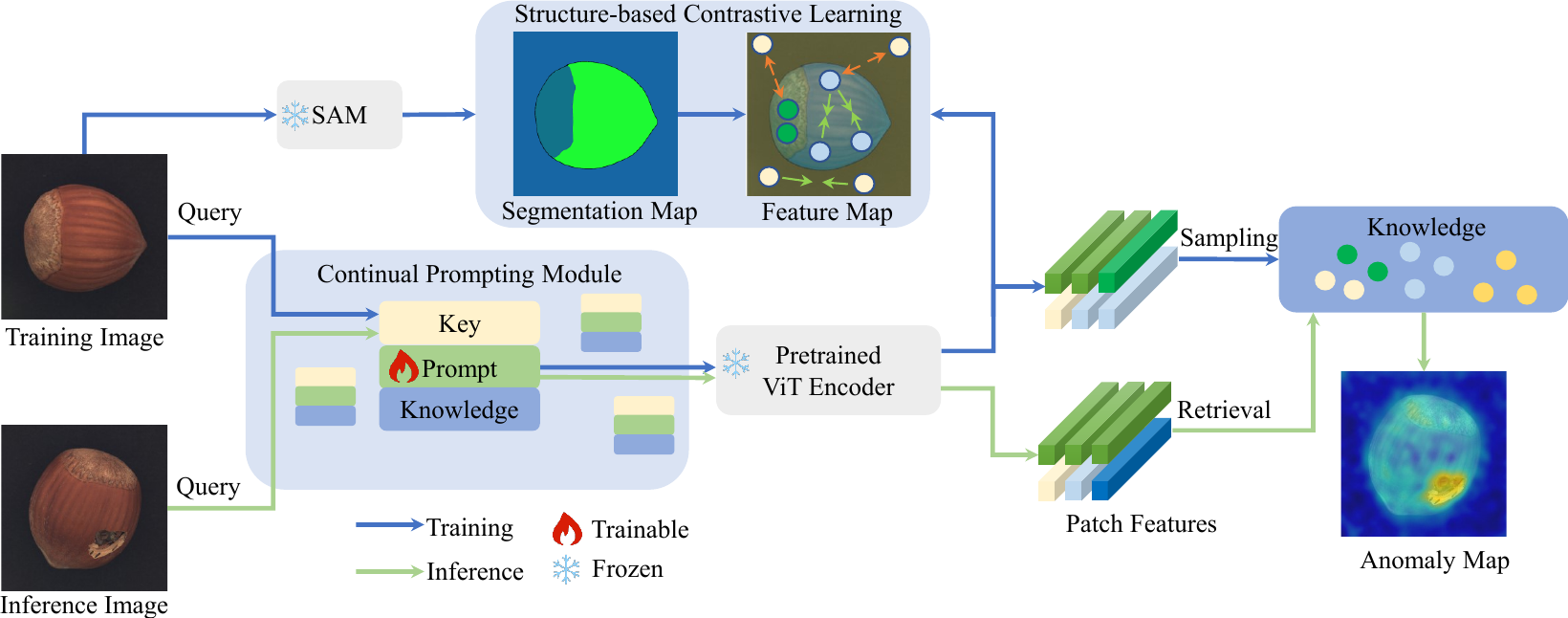} 
\caption{The framework of UCAD mainly comprises a Continual Prompting Module (CPM) and a Structure-based Contrastive Learning (SCL) module, integrated with the SAM network. 
During training,  the CPM establishes a key-prompt-knowledge system that efficiently maintains training data information, while also reducing memory and computational resource usage. 
Moreover, UCAD proposes a contrastive learning method using the SAM segmentation map to enhance the feature representations. 
Finally, the detection of anomalies is accomplished by comparing current features and retrieved task-specific knowledge.}
\label{fig:pipeline}
\end{figure*}

\subsection{Continual Image Anomaly Detection}
Different from natural image object detection tasks, the data stream is common in industrial manufacturing.
Some current methods have recognized this phenomenon and attempted to design algorithms specifically to address the challenges in this scenario. IDDM~\cite{zhang2023iddm} presents an incremental anomaly detection method based on a small number of labeled samples. On the other hand, LeMO~\cite{gao2023towards} follows the common unsupervised anomaly detection paradigm and performs incremental anomaly detection as normal samples continuously increase. However, both IDDM and LeMO focus on intra-class continual anomaly detection research without addressing inter-class incremental anomaly detection challenges. Li et al.'s research~\cite{li2022towards} is the most closely related to ours. They propose DNE for image-level anomaly detection in continual learning scenarios. Due to the limitation of DNE in storing only class-level information, it cannot perform fine-grained localization, thus making it unsuitable for anomaly segmentation. Our method goes beyond continual anomaly classification and extends to pixel-level continual anomaly detection.

\section{Methods}



\subsection{Unsupervised Continual AD Problem Definition}
Unsupervised Anomaly Detection (AD) aims to identify anomalous data using only normal data, since obtaining labeled anomalous samples is challenging in real-world industrial production scenarios.  
The training set contains only normal samples from various tasks, while the test set includes both normal and abnormal samples, reflecting real-world applications.
To formulate the problem, we define the multi-class training set as $\mathcal{T}_{train}^{total} = \left\{\mathcal{T}_{train}^{1}, \mathcal{T}_{train}^{2}, \cdots, \mathcal{T}_{train}^{n} \right\}$ and test set as $\mathcal{T}_{test}^{total} = \left\{\mathcal{T}_{test}^{1}, \mathcal{T}_{test}^{2}, \cdots, \mathcal{T}_{test}^{n} \right\}$. 
$\mathcal{T}_{train}^{i}$ and $\mathcal{T}_{test}^{i}$ represent class $i$-th training and test data, respectively.  

Under unsupervised continual AD and segmentation setting, a unified model is trained non-repetitively on incrementally added classes. Given $N_{task}$ tasks or classes, the model is sequentially trained on sub-training sets $\mathcal{T}_{train}^{i}, i \in N_{task}$ and subsequently tested on all past test subdatasets {$\mathcal{T}_{test}^{total}$}. This evaluation method ensures the final trained model's ability to retain previously acquired knowledge.

\subsection{Continual Prompting Module} \label{sec:CPM}
Applying CL to unsupervised AD faces two challenges: 1) How to determine the task identities of the incoming image automatically; 2) How to guide the model's predictions for the relevant task in an unsupervised manner.
Thus, a continual prompting module is designed to dynamically adapt and instruct unsupervised model predictions. 
We propose to use a memory space $\mathcal{M}$ for a \textbf{key-prompt-knowledge} architecture, $(\mathcal{K}_e, \mathcal{V}, \mathcal{K}_n )$ , that contains two distinct phases: the task identification phase and the task adaptation phase. 

In the task identification phase, images $ x \in \mathbb{R}^{H \times W \times C} $ will go through a frozen pretrained vision transformer $f$ (ViT) to extract keys $k \in \mathcal{K}_e$, also known as task identities. 
Because task identity contains both textual details and high-level information, we use a specific layer of ViT rather than the last embedding $k = f^i(x), k \in \mathbb{R}^{N_p \times C}$, in which $k$ is the feature and $N_p$ is the num of patches after $i$-th block (in this paper, we use $i=5$). 
However, assuming we have $N_I$ training images for task $t$, all extracted embeddings would have dimension 
$\mathcal{K}^t \in \mathbb{R}^{N_I \times N_p \times C}$, which means a lot of memory space. 
To make task matching efficient during testing, we propose to use one image's feature space representing the whole task $\mathbb{R}^{N_I \times N_p \times C} \rightarrow \mathbb{R}^{N_p \times C}$. 
Note that a single image's feature space is negligible compared to the whole task in the continual training setting. 
We find that the farthest point sampling method~\cite{eldar1997farthest} is efficient for selecting representative features to serve as \textbf{keys}. 
So task identities $\mathcal{K}_e$ can be represented as a set:
\begin{equation}
    \begin{aligned}
\mathcal{K}_e^t = FPS(\mathcal{K}^t), \;\;\;\; \mathcal{K}_e^t \in \mathbb{R}^{N_p \times C} \\  
\mathcal{K}_e = \{\mathcal{K}_e^0, \mathcal{K}_e^1, ...,   \mathcal{K}_e^t \}, \;\;\;\; t \in N_{task},
    \end{aligned}
\label{eq:fps_sampling}
\end{equation}
where $FPS$ is furthest point sampling, $\mathcal{K}^t$ represents all extracted embeddings of task $t$.

During the task adaptation phase, inspired by ~\cite{liu2021p} which injects new knowledge into models, we design learnable \textbf{prompts} $\mathcal{V}$ to transfer task-related information to the current image.  
Unlike $\mathcal{K}_e$ that is downsampled from the pretrained backbone, prompts $p \in \mathcal{V}$ are purely learnable to accommodate the current task.  
We add a prompt $p^i$ to each layer's input feature to convey task information to the current image, $k^i = f^i(k^{i-1} + p^i)$, where $k^i$ is the output feature of the $i$-th layer, $k^{i-1}$ is the input feature, and $p^i$ is the prompt added to the $i$-th layer to transfer task-specific information to the current image.
Then, the task-transferred image features $k^i$ are used to create the \textbf{knowledge} $\mathcal{K}_n$ during training. 
Since we are not using supervision, $\mathcal{K}_n$ serves as the standard to distinguish anomaly data by comparing it to the test image features. 
However, image features can be exceedingly large during training accumulation, we use coreset sampling ~\cite{roth2022towards} to reduce storage for $\mathcal{V}$. 
\begin{equation}
    \begin{aligned}
\mathcal{K}_n &= CoreSetSampling(k^i) \\
&= arg \underset{\mathcal{M}_c \subset \mathcal{M}}{\mathrm{min}}\; \underset{m \in \mathcal{M}}{\mathrm{max}} \; \underset{n \in \mathcal{M}_c}{\mathrm{min}} \; ||m - n||_2,
\label{eq:coreset}
\end{aligned}
\end{equation}
 where $\mathcal{M}$ is the nominal image features during training, $\mathcal{M}_c$ is the coreset space for patch-level features $k^i$, and $i=5$ in our experiments since middle features contain both context and semantic information. After establishing key-prompt-knowledge correspondance for each task, our proposed Continual Prompting Module can successfully transfer knowledge from previous tasks to the current image. 
However, the features stored in $\mathcal{V}$ may not be discriminative enough because the backbone $f$ has been pretrained and not adapted to the current task. 
As the original backbone was trained on natural images, we modified it to improve its feature representation for industrial images. Industrial images mainly contain information about texture and edge structure, and the similarity between different industrial product images is often high. This allows us to use fewer features to represent normal industrial images.
To make feature representations more compact, we developed a structure-based contrastive learning method to learn prompt contrastively. 





\subsection{Structure-based Contrastive Learning}

Inspired by ReConPatch~\cite{hyun2023reconpatch}, we designed structure-based contrastive learning to enhance the network representation for patch-level comparison during testing. 
We discovered that SAM~\cite{Kirillov_2023_ICCV} consistently provides general structure knowledge, such as masks, without requiring training.
As illustrated in Figure~\ref{fig:pipeline}, for each image in the training set, we employed SAM to generate corresponding segmentation images $I_s$, in which different regions represent distinct structures or semantics. 
Simultaneously, guided by prompts, we obtain the feature map $F_s \in k^i$ for each region, where $k^i$ is the $i$-th layer feature in the previous section. 
We downsampled the segmentation image $I_s$ to match the size of $F_s \in \mathbb{R}^{c \times h \times w}$ and aligned the corresponding positions to create the label map $L_s$. 
By incorporating contrastive learning, the knowledge generality in $\mathcal{K}_n$ is achieved by pulling the features of the same region closer and pushing the features of different regions further apart. The loss function is:
\begin{equation}
    \begin{aligned}
L_{pos\_con} &= \sum\limits_{i,p}^{H}{\sum\limits_{j,q}^{W}{cos(F_{ij},F{p,q})}},(L_{ij}=L_{pq}), \\
L_{neg\_con} &= \sum\limits_{i,p}^{H}{\sum\limits_{j,q}^{W}{cos(F_{ij},F{p,q})}},(L_{ij}\ne L_{pq}), \\
L_{total} &= \lambda_{\alpha} L_{neg\_con}-\lambda_{\beta}L_{pos\_con}. \\
\label{eq:con_loss}
\end{aligned}
\end{equation}
In the given paragraph, $F_{ij}$ denotes the embedding of feature $F_s$ at position $(i, j)$ with a shape of $(1, 1, c)$, while $L_{ij}$ represents the label of feature $F_{ij}$ at the corresponding position in the segmentation result generated by SAM. $\lambda_{\alpha}$ and $\lambda_{\beta}$ are 1.
By training prompts using this contrastive loss, the model's representation ability is enhanced, and features of various textures become more compact. 
Consequently, this approach results in more distinct representations of abnormal features during testing, allowing them to stand out prominently.


\subsection{Test-Time Task-Agnostic Inference}
\subsubsection{Task Selection and Adaption}
To automatically determine the task identity during testing, an image $x^{test}$ initially locates its corresponding task by selecting from $\mathcal{K}_e$ based on the highest similarity. 
The corresponding task identity is selected by the equation below:
\begin{equation}
\begin{aligned}
\mathcal{K}_e^t &= \underset{m \in \mathcal{K}_e}{arg \; min} \; Sim(m - m^{test}), \\
Sim(m - m^{test}) &=\sum_{x \in N_p}^{} \underset{y \in N_p}{min} || m_x - m^{test}_y ||_2,
\label{eq:patch_retrieval}
\end{aligned}
\end{equation}
where $m^{test}$ is the patch-level feature from $i$-th layer feature map of ViT containing multiple patches $N_p$, $i = 5$ in this paper as discussed in previous section. 
Since the utilization of a key-prompt-knowledge architecture, the associated prompts $\mathcal{V}$ and knowledge $\mathcal{K}_n$ can be readily retrieved. 
By combining the selected prompts with test patches and processing them through ViT, features from the test sample are adapted and extracted. 
Subsequently, anomaly scores are calculated based on the minimum distance to the task's knowledge $\mathcal{K}_n^t$.




\begin{table*}[tb]
\centering
\resizebox{\textwidth}{!}{
\begin{tabular}{lccccccccccccccccc}
\hline
Methods             & Bottle & cable & capsule & carpet & grid  & hazelnut & leather & metal\_nut & pill  & screw & tile  & toothbrush & transistor & wood  & zipper & average & avg FM    \\ \hline
CFA                 & 0.309  & 0.489 & 0.275   & 0.834  & 0.571 & 0.903    & 0.935   & 0.464      & 0.528 & 0.528 & 0.763 & 0.519      & 0.320      & 0.923 & 0.984  & 0.623   & 0.361 \\
CSFlow             & 0.129  & 0.420 & 0.363   & 0.978  & 0.602 & 0.269    & 0.906   & 0.220      & 0.263 & 0.434 & 0.697 & 0.569      & 0.432      & 0.802 & 0.997  & 0.539   & 0.426 \\
CutPaste           & 0.111  & 0.422 & 0.373   & 0.198  & 0.214 & 0.578    & 0.007   & 0.517      & 0.371 & 0.356 & 0.112 & 0.158      & 0.340      & 0.150 & 0.775  & 0.312   & 0.510 \\
DRAEM               & 0.793  & 0.411 & 0.517   & 0.537  & 0.799 & 0.524    & 0.480   & 0.422      & 0.452 & \textbf{1.000} & 0.548 & 0.625      & 0.307      & 0.517 & \textbf{0.996}  & 0.595   & 0.371 \\
FastFlow            & 0.454  & 0.512 & 0.517   & 0.489  & 0.482 & 0.522    & 0.487   & 0.476      & 0.575 & 0.402 & 0.489 & 0.267      & 0.526      & 0.616 & 0.867  & 0.512   & 0.279 \\
FAVAE               & 0.666  & 0.396 & 0.357   & 0.610  & 0.644 & 0.884    & 0.406   & 0.416      & 0.531 & 0.624 & 0.563 & 0.503      & 0.331      & 0.728 & 0.544  & 0.547   & 0.102 \\
PaDiM               & 0.458  & 0.544 & 0.418   & 0.454  & 0.704 & 0.635    & 0.418   & 0.446      & 0.449 & 0.578 & 0.581 & 0.678      & 0.407      & 0.549 & 0.855  & 0.545   & 0.368 \\
PatchCore           & 0.163  & 0.518 & 0.350   & 0.968  & 0.700 & 0.839    & 0.625   & 0.259      & 0.459 & 0.484 & 0.776 & 0.586      & 0.341      & 0.970 & 0.991  & 0.602   & 0.383 \\
RD4AD               & 0.401  & 0.538 & 0.475   & 0.583  & 0.558 & 0.909    & 0.596   & 0.623      & 0.479 & 0.596 & 0.715 & 0.397      & 0.385      & 0.700 & 0.987  & 0.596   & 0.393 \\
SPADE               & 0.302  & 0.444 & 0.525   & 0.529  & 0.460 & 0.410    & 0.577   & 0.592      & 0.484 & 0.514 & 0.881 & 0.386      & 0.622      & 0.897 & 0.949  & 0.571   & 0.285 \\
STPM                & 0.329  & 0.539 & 0.610   & 0.462  & 0.569 & 0.540    & 0.740   & 0.456      & 0.523 & 0.753 & 0.736 & 0.375      & 0.450      & 0.779 & 0.783  & 0.576   & 0.325 \\
SimpleNet           & 0.938  & 0.560 & 0.519   & 0.736  & 0.592 & 0.859    & 0.749   & 0.710      & 0.701 & 0.599 & 0.654 & 0.422      & 0.669      & 0.908 & 0.996  & 0.708   & 0.211 \\
UniAD               & 0.801  & 0.660 & 0.823   & 0.754  & 0.713 & 0.904    & 0.715   & 0.791      & 0.869 & 0.731 & 0.687 & 0.776      & 0.490      & 0.903 & 0.997  & 0.774   & 0.229 \\
\hline
DNE                 & 0.990  & 0.619 & 0.609   & 0.984  & \textbf{0.998} & 0.924    & \textbf{1.000}   & \textbf{0.989}      & 0.671 & 0.588 & 0.980 & 0.933      & 0.877      & 0.930 & 0.958  & 0.870   & 0.116 \\
PatchCore* & 0.533  & 0.505 & 0.351   & 0.865  & 0.723 & 0.959    & 0.854   & 0.456      & 0.511 & 0.626 & 0.748 & 0.600      & 0.427      & 0.900 & 0.974  & 0.669   & 0.318 \\
UniAD*     & 0.997  & 0.701 & 0.765   & \textbf{0.998}  & 0.896 & 0.936    & \textbf{1.000}   & 0.964      & \textbf{0.895} & 0.554 & 0.989 & 0.928      & \textbf{0.966}      & 0.982 & 0.987  & 0.904   & 0.076 \\
\hline
Ours                & \textbf{1.000} &	\textbf{0.751} &	\textbf{0.866} &	0.965 &	0.944 &	\textbf{0.994} &	\textbf{1.000} &	0.988 &	0.894 &	0.739 &	\textbf{0.998} &	\textbf{1.000} &	0.874 &	0.995 &	0.938 &	\textbf{0.930} &	\textbf{0.010}  \\ \hline
\end{tabular}}
\caption{Image-level AUROC$\uparrow$ and corrsponding FM$\downarrow$ on MVTec AD dataset~\cite{bergmann2019mvtec} after training on the last subdataset. Note that * signifies the usage of a cache pool for rehearsal during training which may not be possible in real applications. The best results are highlighted in bold.}
\label{tab:image_mvtec}
\end{table*}

\begin{table*}[]
\centering
\resizebox{\textwidth}{!}{
\begin{tabular}{lccccccccccccccccc}
\hline
Methods             & Bottle & cable & capsule & carpet & grid  & hazelnut & leather & metal\_nut & pill  & screw & tile  & toothbrush & transistor & wood  & zipper & average & avg FM    \\ \hline
CFA                 & 0.068 & 0.056 & 0.050 & 0.271 & 0.004 & 0.341 & \textbf{0.393} & 0.255 & 0.080 & 0.015 & 0.155 & 0.053 & 0.056 & 0.281 & 0.573 & 0.177 & 0.083 \\
DRAEM               & 0.117 & 0.019 & 0.044 & 0.018 & 0.005 & 0.036 & 0.013 & 0.142 & 0.104 & 0.002 & 0.130 & 0.039 & 0.040 & 0.033 & \textbf{0.734} & 0.098 & 0.116 \\
FastFlow            & 0.044 & 0.021 & 0.013 & 0.013 & 0.005 & 0.028 & 0.007 & 0.090 & 0.029 & 0.003 & 0.060 & 0.015 & 0.036 & 0.037 & 0.264 & 0.044 & 0.214 \\
FAVAE               & 0.086 & 0.048 & 0.039 & 0.015 & 0.004 & 0.389 & 0.112 & 0.174 & 0.070 & 0.017 & 0.064 & 0.043 & 0.046 & 0.093 & 0.039 & 0.083 & 0.083 \\
PaDiM               & 0.072 & 0.037 & 0.030 & 0.023 & 0.006 & 0.183 & 0.039 & 0.155 & 0.044 & 0.014 & 0.065 & 0.044 & 0.049 & 0.080 & 0.452 & 0.086 & 0.366 \\
PatchCore           & 0.048 & 0.029 & 0.035 & 0.552 & 0.003 & 0.338 & 0.279 & 0.248 & 0.051 & 0.008 & 0.249 & 0.034 & 0.079 & 0.304 & 0.595 & 0.190 & 0.371 \\
RD4AD               & 0.055 & 0.040 & 0.064 & 0.212 & 0.005 & 0.384 & 0.116 & 0.247 & 0.061 & 0.015 & 0.193 & 0.034 & 0.059 & 0.097 & 0.562 & 0.143 & 0.425 \\
SPADE               & 0.122 & 0.052 & 0.044 & 0.117 & 0.004 & 0.512 & 0.264 & 0.181 & 0.060 & 0.020 & 0.096 & 0.043 & 0.050 & 0.172 & 0.531 & 0.151 & 0.319 \\
STPM                & 0.074 & 0.019 & 0.073 & 0.054 & 0.005 & 0.037 & 0.108 & 0.354 & 0.111 & 0.001 & 0.397 & 0.046 & 0.046 & 0.119 & 0.203 & 0.110 & 0.352 \\
SimpleNet           & 0.108 & 0.045 & 0.029 & 0.018 & 0.004 & 0.029 & 0.006 & 0.227 & 0.077 & 0.004 & 0.082 & 0.046 & 0.049 & 0.037 & 0.139 & 0.060 & 0.069 \\
UniAD               & 0.054 & 0.031 & 0.022 & 0.047 & 0.007 & 0.189 & 0.053 & 0.110 & 0.034 & 0.008 & 0.107 & 0.040 & 0.045 & 0.103 & 0.444 & 0.086 & 0.419 \\
\hline
PatchCore* & 0.087 & 0.043 & 0.042 & 0.407 & 0.003 & 0.443 & 0.352 & 0.189 & 0.058 & 0.017 & 0.124 & 0.028 & 0.053 & 0.270 & 0.604 & 0.181 & 0.343 \\
UniAD*     & 0.734 & 0.232 & 0.313 & 0.517 & \textbf{0.204} & 0.378 & 0.360 & 0.587 & 0.346 & 0.035 & 0.428 & \textbf{0.398} & \textbf{0.542} & 0.378 & 0.443 & 0.393 & 0.086 \\
\hline
Ours       & \textbf{0.752} &	\textbf{0.290} &	\textbf{0.349} &	\textbf{0.622} &	0.187 &	\textbf{0.506} &	0.333 &	\textbf{0.775} &	\textbf{0.634} &	\textbf{0.214} &	\textbf{0.549} &	0.298 &	0.398 &	\textbf{0.535} &	0.398 &	\textbf{0.456} &	\textbf{0.013} 
   \\ \hline
\end{tabular}}
\caption{Pixel-level AUPR$\uparrow$ and corrsponding FM$\downarrow$ on MVTec AD dataset~\cite{bergmann2019mvtec} after training on the last subdataset.}
\label{tab:pixel_mvtec}
\end{table*}

\subsubsection{Anomaly Detection and Segmentation}
To calculate the anomaly score, we compare the image feature $m^{test}$ with the nominal features stored in task-specific knowledge base $\mathcal{K}_n^t$.
Building upon the patch-level retrieval, we employed re-weighting to implement the anomaly detection process. $\mathcal{N}_{b}(m^{*})$ represents the nearest neighbors of $m^*$ in $\mathcal{K}_n^t$. 
We use the distance between $m^{test}$ and $m^*$ as the basic anomaly score, and then calculate the distance between $m^{test}$ and the features in $\mathcal{N}_{b}(m^{*})$ to achieve the re-weighting effect. 
Through Eqution~\ref{eq:find_mtest}, we set the furthest distance between feature $m^{test,*}$ in the test feature set $\mathcal{P}(x^{test})$ and memory bank $\mathcal{K}_n^r$ to represent the anomaly score $s^*$ of the sample. 
\begin{equation}
\begin{aligned}
m^{test,*},m^{*} &= \argmax \limits_{m^{test}\in \mathcal{P}(x^{test})} \argmin \limits_{m \in \mathcal{K}_n^t}\left\| m^{test} - m^{l}\right\|_{2}, \\
\quad s^{*} &= \left\| m^{test,*} - m^{*}\right\|_{2}. \\
\end{aligned}
\label{eq:find_mtest}
\end{equation}

By re-weighting from neighbors $m* \in \mathcal{K}_n^t$, the anomaly score $s$ becomes more robust, as in Equation~\ref{eq:calc_score}:
\begin{equation}
    s = \left ( 1- \frac{\mathrm{exp}\left\|m^{test, *} - m^{*} \right\|_{2}}{\sum_{m \in \mathcal{N}_{b}(m^{*})}\mathrm{exp} \left\|m^{test, *} - m \right\|_{2}} \right ) \cdot s^{*}.
\label{eq:calc_score}
\end{equation}
The anomaly score of the whole image is calculated by the max score of all patches, $S_{img} = max(s_i), i \in N_p$. The coarse segmentation map, $S_{cmap}$, is represented by scores calculated from each patch. By upsampling and applying Gaussian smoothing to $S_{cmap}$, the final segmentation result $S_{map}$ is obtained with the same dimensions as the input image.

\begin{table*}[tb]
\centering
\resizebox{\textwidth}{!}{
\begin{tabular}{lcccccccccccccc}
\hline
Methods             & candle & capsules & cashew & chewinggum & fryum & macaroni1 & macaroni2 & pcb1  & pcb2  & pcb3  & pcb4  & pipe\_fryum & average & avg FM    \\ \hline
CFA                 & 0.512  & 0.672    & 0.873  & 0.753      & 0.304 & 0.557     & 0.422     & 0.698 & 0.472 & 0.449 & 0.407 & 0.998       & 0.593   & 0.327 \\
RD4AD               & 0.380  & 0.385    & 0.737  & 0.539      & 0.533 & 0.607     & 0.487     & 0.437 & 0.672 & 0.343 & 0.187 & \textbf{0.999}       & 0.525   & 0.423 \\
PatchCore           & 0.401  & 0.605    & 0.624  & 0.907      & 0.334 & 0.538     & 0.437     & 0.527 & 0.597 & 0.507 & 0.588 & 0.998       & 0.589   & 0.361 \\
SimpleNet           & 0.504  & 0.474    & 0.794  & 0.721      & 0.684 & 0.567     & 0.447     & 0.598 & 0.629 & 0.538 & 0.493 & 0.945       & 0.616   & 0.283 \\
UniAD               & 0.573  & 0.599    & 0.661  & 0.758      & 0.504 & 0.559     & 0.644     & 0.749 & 0.523 & 0.547 & 0.562 & 0.989       & 0.639   & 0.297 \\
\hline
DNE                 & 0.486  & 0.413    & 0.735  & 0.585      & 0.691 & 0.584     & 0.546     & 0.633 & 0.693 & 0.642 & 0.562 & 0.747       & 0.610   & 0.179 \\
PatchCore* & 0.647  & 0.579    & 0.669  & 0.735      & 0.431 & 0.631     & 0.624     & 0.617 & 0.534 & 0.479 & 0.645 & \textbf{0.999}       & 0.633   & 0.349 \\
UniAD*     & \textbf{0.884}  & 0.669    & 0.938  & \textbf{0.970}      & 0.812 & 0.753     & 0.570     & 0.872 & 0.766 & 0.708 & \textbf{0.967} & 0.990       & 0.825   & 0.125 \\
\hline
Ours                & 0.778 	& \textbf{0.877} & 	\textbf{0.960} & 	0.958 & 	\textbf{0.945} &	\textbf{0.823} &	\textbf{0.667} &	\textbf{0.905} &	\textbf{0.871} &	\textbf{0.813} &	0.901 &	0.988 &	\textbf{0.874} &	\textbf{0.039}  \\ \hline
\end{tabular}}
\caption{Image-level AUROC$\uparrow$ and corrsponding FM$\downarrow$ on VisA dataset~\cite{zou2022spot} after training on the last subdataset.}
\label{tab:image_visa}
\end{table*}

\begin{table*}[]
\centering
\resizebox{\textwidth}{!}{
\begin{tabular}{lcccccccccccccc}
\hline
Methods             & candle & capsules & cashew & chewinggum & fryum & macaroni1 & macaroni2 & pcb1  & pcb2  & pcb3  & pcb4  & pipe\_fryum & average & avg FM    \\ \hline
CFA                 & 0.017  & 0.005    & 0.059  & 0.243      & 0.085 & 0.001     & 0.001     & 0.013 & 0.006 & 0.008 & 0.015 & \textbf{0.592}       & 0.087   & 0.184 \\
RD4AD               & 0.002  & 0.005    & 0.061  & 0.045      & 0.098 & 0.001     & 0.001     & 0.013 & 0.008 & 0.008 & 0.013 & 0.576       & 0.069   & 0.201 \\
PatchCore           & 0.012  & 0.007    & 0.055  & 0.315      & 0.082 & 0.000     & 0.000     & 0.008 & 0.004 & 0.007 & 0.010 & 0.585       & 0.090   & 0.311 \\
SimpleNet           & 0.001  & 0.004    & 0.017  & 0.007      & 0.047 & 0.000     & 0.000     & 0.013 & 0.003 & 0.004 & 0.009 & 0.058       & 0.014   & 0.016 \\
UniAD               & 0.006  & 0.013    & 0.040  & 0.185      & 0.087 & 0.002     & 0.002     & 0.015 & 0.005 & 0.015 & 0.013 & 0.576       & 0.080   & 0.218 \\
\hline
PatchCore*  & 0.018  & 0.010    & 0.047  & 0.202      & 0.081 & 0.003     & 0.001     & 0.008 & 0.004 & 0.008 & 0.010 & 0.443       & 0.070   & 0.327 \\
UniAD*     & \textbf{0.132}  & 0.123    & 0.378  & \textbf{0.574}      & \textbf{0.404} & \textbf{0.041}     & \textbf{0.010}     & 0.612 & 0.083 & \textbf{0.266} & \textbf{0.232} & 0.549       & 0.283   & 0.062 \\
\hline
Ours                & 0.067 &	\textbf{0.437} &	\textbf{0.580} &	0.503 &	0.334 &	0.013 &	0.003 &	\textbf{0.702} &	\textbf{0.136} &	\textbf{0.266} &	0.106 &	0.457 &	\textbf{0.300} &	\textbf{0.015} 
   \\ \hline
\end{tabular}}
\caption{Pixel-level AUPR$\uparrow$ and corrsponding FM$\downarrow$ on VisA dataset~\cite{zou2022spot} after training on the last subdataset.}
\label{tab:pixel_visa}
\end{table*}

\section{Experiments and Discussion}

\subsection{Experiments setup}

\textbf{Datasets}
MVTec AD~\cite{bergmann2019mvtec} is the most widely used dataset for industrial image anomaly detection. VisA~\cite{zou2022spot} is now the largest dataset for real-world industrial anomaly detection with pixel-level annotations. We conduct experiments on these two datasets.


\textbf{Methods}
Based on the anomaly methods discussed in our related work section and previous benchmark~\cite{xie2023iad}, we selected the most representative methods from each paradigm to establish the benchmark. These methods include 
CFA~\cite{lee2022cfa}, CSFlow~\cite{rudolph2022fully}, CutPaste~\cite{li2021cutpaste}, DNE~\cite{li2022towards}, DRAEM~\cite{zavrtanik2021draem}, FastFlow~\cite{yu2021fastflow}, FAVAE~\cite{dehaene2020anomaly}, PaDiM~\cite{defard2021padim}, PatchCore~\cite{roth2022towards}, RD4AD~\cite{deng2022anomaly}, SPADE~\cite{cohen2020sub}, STPM~\cite{Wang2021StudentTeacherFP}, SimpleNet~\cite{liu2023simplenet}, and UniAD~\cite{you2022unified}.

\textbf{Metrics}
Following the common practice, we utilize Area Under the Receiver Operating Characteristics (AU-ROC/AUC) to assess the model's ability in anomaly classification. For pixel-level anomaly segmentation capability, we employ Area Under Precision-Recall (AUPR/AP) for model evaluation. In addition, we use Forgetting Measure (FM)~\cite{chaudhry2018riemannian} to evaluate models' ability to prevent catastrophic forgetting. 


\begin{equation}
avg\ FM = \frac{1}{k-1}\sum \limits_{j=1}^{k-1} \max\limits_{l\in\{1,...,k-1\}}\mathbf{T}_{l,j}-\mathbf{T}_{k,j},
\label{eq:avg_fm}
\end{equation}
where $\mathbf{T}$ represents tasks, $k$ stands for the current training task ID, and $j$ refers to the task ID being evaluated. And $avg\ FM$ represents the average forgetting measure of the model after completing $k$ tasks. During the inference, we evaluate the model after training on all tasks.

\textbf{Training Details and Module Parameter Settings} We utilized the \textit{vit-base-patch16-224} backbone pretrained on ImageNet 21K~\cite{Deng2009ImageNetAL} for our method. During prompt training, we employed a batch size of 8 and adapt Adam optimizer~\cite{Kingma2014AdamAM} with a learning rate of 0.0005 and momentum of 0.9. The training process spanned 25 epochs. Our key-prompt-knowledge structure comprised a key of size (15, 196, 1024) float array, a prompt of size (15, 7, 768) float array, and knowledge of size (15, 196, 1024) float array, with an overall size of approximately 23.28MB. 

\subsection{Continual anomaly detection benchmark}
We conducted comprehensive evaluations of the aforementioned 14 methods on the MVTec AD and VisA datasets. 
Among them, DNE stands as the SOTA method in unsupervised continual AD. 
Meanwhile, PatchCore and UniAD are two representative AD methods for memory-based and unified methods, respectively. 
Intuitively, these two methods appear to be better suited for the continual learning scenario. 
Due to the famous replay in continual learning methods, we also conducted replay-based experiments on PatchCore and UniAD. In these experiments, we provided them with a buffer capable of storing 100 training samples.

\textbf{Quantitative Analysis} As shown in Tables~\ref{tab:image_mvtec} -~\ref{tab:pixel_visa}, most of the anomaly detection methods experienced significant performance degradation in the context of continual learning scenarios. Surprisingly, with the use of replay, UniAD managed to surpass DNE on the MVTec AD dataset. Moreover, on the VisA dataset, even without replay, UniAD outperformed DNE. On the other hand, our method achieved a substantial lead over the second-best approach without the use of replay. Specifically, on the MVTec AD dataset, our method shows a 2.6 point lead in Image AUROC and a 6.3 point lead in Pixel AUPR over the second-ranked method, while on the VisA dataset, we achieve a 4.9 point lead in Image AUROC and a 1.7 point lead in Pixel AUPR. It can be observed that on the more complex structural VisA dataset, the detection capability of DNE, which solely relies on class tokens for anomaly discrimination, is significantly reduced. In contrast, our method remains unaffected.

Based on the comprehensive experimental results, our approach shows significant improvement over other methods in detecting anomalies under a continual setting. 
The experiments also demonstrate the potential of reconstruction-based methods, such as UniAD, in the field of continual UAD.
In future works, combining our suggested CPM with the reconstruction-based UAD approach could be beneficial.

\begin{figure*}[t]
\centering
\includegraphics[width=1.95\columnwidth]{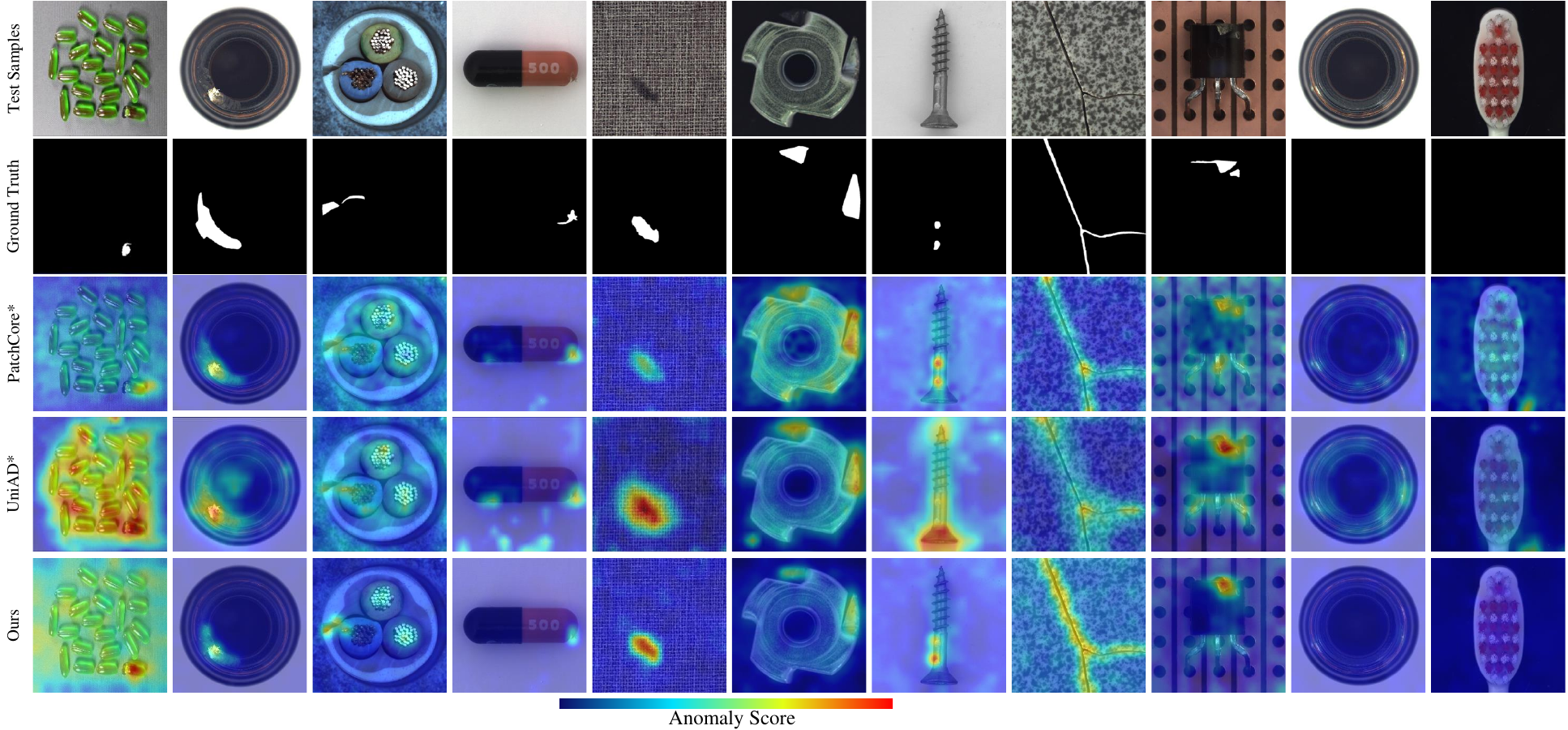} 
\caption{Visualization examples of continual anomaly detection. The first row displays the original anomaly images, the second row shows the ground truth annotations, and the third to fifth rows depict the heatmaps of our method and other methods.}
\label{fig:vis_comparsion}
\end{figure*}

\textbf{Qualitative Analysis} As illustrated in Figure~\ref{fig:vis_comparsion}, our method demonstrates the ability to roughly predict the locations of anomalies. This progress stands as a significant improvement compared to DNE. Compared to PatchCore* and UniAD*, our method exhibits two distinct advantages. Firstly, it demonstrates a more precise localization of anomalies. Secondly, it minimizes false positives in normal image classification.

\subsection{Ablation study}


\begin{table}[]
\centering
\small
\begin{tabular}{cccc}
\hline
CPM & SCL & MVTec AD & VisA  \\ 
\hline
    \ding{55}        &      \ding{55}      & 0.693/0.183                  & 0.584/0.050              \\
\checkmark            &       \ding{55}                 & 0.894/0.426                  & 0.786/0.251              \\
\checkmark            & \checkmark               & \textbf{0.930/0.456}                  & \textbf{0.874/0.300 }       \\ \hline
\end{tabular}
\caption{Ablation study for CPM and SCL.}
\label{tab:abliation_cpm}
\end{table}

\textbf{Module Effectivity} As shown in Table~\ref{tab:abliation_cpm}, We analyze the impact of two modules - Continual Prompting Module (CPM) and Structure-based Contrastive Learning (SCL). We observed significant improvements in the model's performance with the implementation of these modules. 
In the absence of CPM's key-prompt-knowledge architecture,  our model used a single Knowledge base and reset it every time a new task was introduced. This approach restricted the model's ability to adapt to continual learning without supervision. 
However, with the inclusion of CPM, the model's Image AUROC score showed a significant improvement of 20 points.
Regarding SCL, we found that without learning prompts contrastively, the model relied solely on the frozen ViT for feature extraction. This approach leds to a drop of around 4 points in the final performance, indicating the importance of SCL's feature generalizability improvement.


\begin{table}[]
\centering
\small
\begin{tabular}{cccccc}
\hline
\multirow{2}{*}{CPM} & \multirow{2}{*}{SCL} & Knowledge & MVTec AD & VisA   \\
                     &                      & Size      & Metric   & Metric \\ \hline
\checkmark           &    \ding{55}        & 1x        & 0.894/0.426                  & 0.786/0.251              \\
\checkmark           &     \ding{55}       & 2x        & 0.921/0.452                  & 0.818/0.255              \\
\checkmark           &     \ding{55}       & 4x        &  0.929/0.453             &
0.860/0.294              \\
\checkmark           & \checkmark           & 1x          & 0.930/0.456                  & 0.874/0.300              \\
\checkmark           & \checkmark           & 2x        & 0.936/0.461                  & 0.893/0.307              \\
\checkmark           & \checkmark           & 4x        & \textbf{0.938}/\textbf{0.466}                  & \textbf{0.909}/\textbf{0.310}              \\
 \hline
\end{tabular}
\caption{Ablation study for \textit{Knowledge} size and SCL.}
\label{tab:abliation_knowledge}
\end{table}


\begin{table}[t]
\centering
\small
\begin{tabular}{cccccc}
\hline
 Encoder & MVTec AD & VisA   \\
 Layer   & Metric   & Metric \\ \hline
 1             & 0.840/0.399                  & 0.806/0.143              \\
 3             & 0.934/0.451                  & \textbf{0.876}/0.283              \\
 5             & 0.930/\textbf{0.456}                  & 0.874\textbf{/0.300}              \\
 7             & \textbf{0.936}/0.444                  & 0.872/0.267              \\
 9             & 0.906/0.420                  & 0.853/0.248              \\      
 \hline
\end{tabular}
\caption{Ablation study for ViT encoder layer.}
\label{tab:abliation_layer}
\end{table}

\textbf{Size of Knowledge Base in CPM} To further investigate the role of SCL, we designed ablation experiments as illustrated in Table~\ref{tab:abliation_knowledge}, by altering the size of Knowledge within the CPM module. 
The basic \textit{Knowledge} size is 196, corresponding to the number of patches in a single image.
Our method enables the representation of all images' patches in a task with a single image feature space. 
Intriguingly, in the presence of SCL, even when the \textit{Knowledge} size is 4 times larger, the performance enhancement remains marginal.
However, without SCL, as the \textit{Knowledge} size increases, the model exhibits a noticeable performance gain. 
This phenomenon can be attributed to SCL's capacity to render feature distributions more compact, allowing a feature of the same size to encapsulate additional information.

\textbf{ViT Feature Layers} Furthermore, we explore the number of layers to use from ViT encoder in our method. The results in Table~\ref{tab:abliation_layer} indicate that neither shallow nor deep layers are effective for unsupervised anomaly detection. 
Intermediate layers, on the other hand, perform better as they are capable of representing both contextual and semantic information.
We found that various datasets possess nuances in their definitions of anomalies, resulting in varying levels of granularity.
While the degree of contextual knowledge required may vary across different datasets, we decided to stick with the fifth layer for simplicity.

\section{Conclusion}
In this paper, we investigate the problem of applying continual learning on unsupervised anomaly detection to address real-world applications in industrial manufacturing. 
To facilitate this research, we build a comprehensive benchmark for unsupervised continual anomaly detection and segmentation. Furthermore, our proposed UCAD for task-agnostic CL on UAD is the first study to design a pixel-level continual anomaly segmentation method. UCAD novelty relies on a continual prompting module and structured-based contrastive learning, significantly improving continual anomaly detection performance. 
Comprehensive experiments have underscored our framework's efficacy and robustness with varying hyperparameters.  
We also find that amalgamating and prompting ViT features from various layers might further enhance results, which we leave for future endeavors.

\appendix

\section{Acknowledgments}
This work is supported by the National Key R\&D Program of China (Grant NO. 2022YFF1202903) and the National Natural Science Foundation of China (Grant NO. 62122035).


\bibliography{aaai_ucad}

\clearpage

\section{Appendix}

\subsection{Dataset}

MVTec AD~\cite{bergmann2019mvtec} is the most widely used dataset for industrial image anomaly detection, and it comprises of 15 categories of items, comprising a collection of 1725 normal and abnormal photos, as well as a total of 3629 normal images, as a training set. The resolution of each image ranges from 700$\times$700 to 1024$\times$1024 pixels.

VisA~\cite{zou2022spot} is now the largest dataset for real-world industrial anomaly detection with pixel-level annotations. The VisA dataset is divided into twelve categories. There are 9,621 normal samples and 1,200 abnormal samples in 10,821 images. The abnormal images have both structural problems, like parts that are out of place or missing, and surface problems, like scratches, dents, or cracks.

\subsection{Visualization}

\begin{figure*}[b]
\centering
\includegraphics[width=2.8\columnwidth]{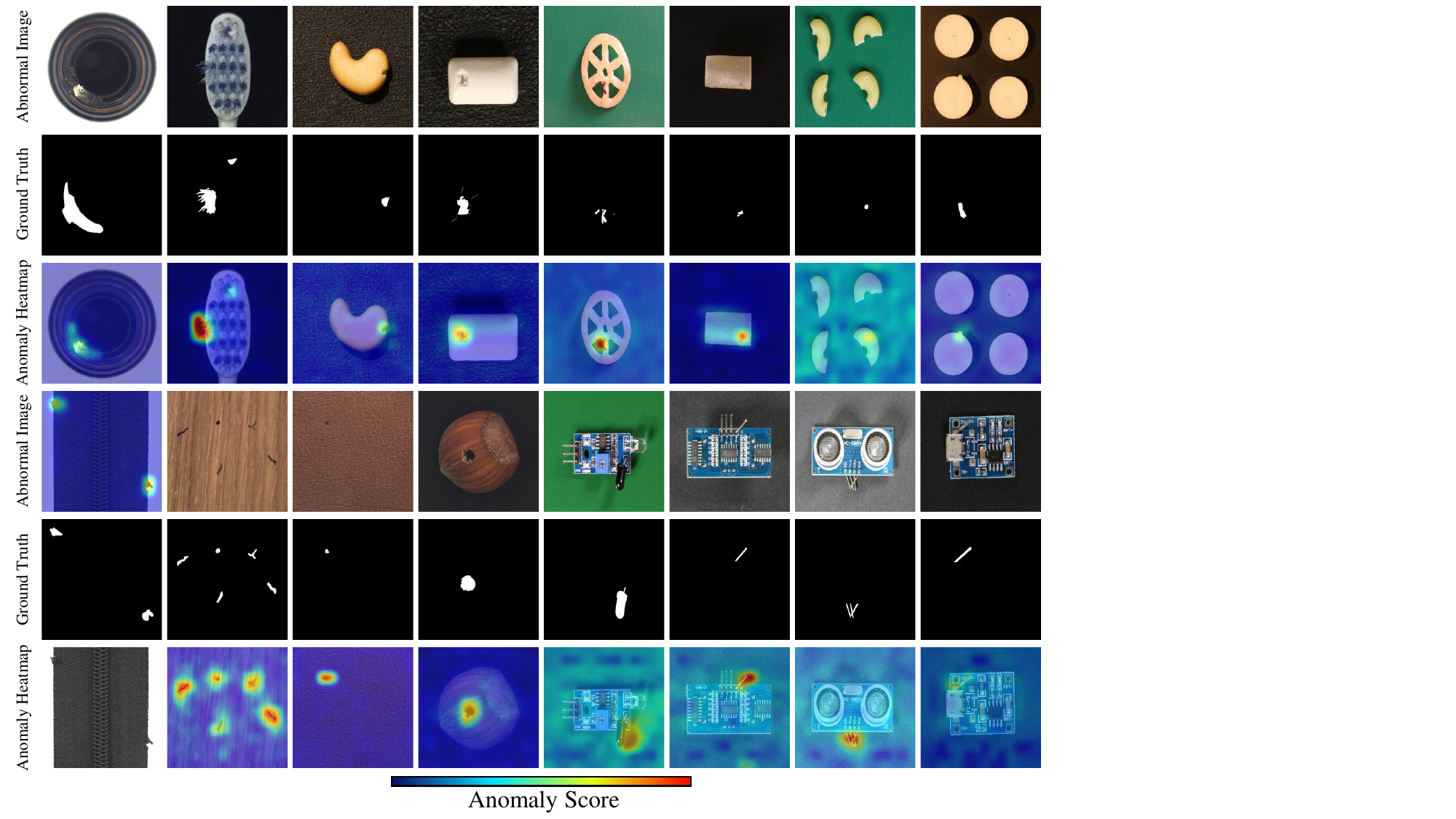} 
\caption{Visualization examples of continual anomaly detection. The first row displays the original anomaly images, the second row shows the ground truth annotations, and the third to fifth rows depict the heatmaps of our method and other methods.}
\label{fig:vis_comparsion_supp}
\end{figure*}

Here, we provide more examples of result visualization in Figure~\ref{fig:vis_comparsion_supp}.

\subsection{Methods}
As discussed in the main text, we select the most representative methods from each paradigm to establish the benchmark. These methods include 
CFA~\cite{lee2022cfa}, CSFlow~\cite{rudolph2022fully}, CutPaste~\cite{li2021cutpaste}, DNE~\cite{li2022towards}, DRAEM~\cite{zavrtanik2021draem}, FastFlow~\cite{yu2021fastflow}, FAVAE~\cite{dehaene2020anomaly}, PaDiM~\cite{defard2021padim}, PatchCore~\cite{roth2022towards}, RD4AD~\cite{deng2022anomaly}, SPADE~\cite{cohen2020sub}, STPM~\cite{Wang2021StudentTeacherFP}, SimpleNet~\cite{liu2023simplenet}, and UniAD~\cite{you2022unified}.

\begin{table*}[!t]
\resizebox{\textwidth}{!}{
\begin{tabular}{lccccccccccccccc}
\hline
               & CFA   & CSFlow & CutPaste & DNE    & DRAEM  & FastFlow & FAVAE   & PaDiM            & PatchCore        & RD4AD & SPADE            & STPM & SimpleNet & UniAD  & Ours    \\ \hline
training epoch & 50    & 240    & 256      & 50     & 700    & 500      & 100     & 1                & 1                & 200   & 1                & 100  & 40        & 1000   & 25      \\
batch size     & 4     & 16     & 32       & 32     & 8      & 32       & 64      & 32               & 2                & 8     & 8                & 8    & 8         & 32     & 8       \\
image size     & 256   & 768    & 224      & 224    & 256    & 256      & 256     & 256              & 256              & 256   & 256              & 256  & 256       & 224    & 224     \\
learning rate  & 0.001 & 0.0002 & 0.0001   & 0.0001 & 0.0001 & 0.001    & 0.00001 & \textbackslash{} & \textbackslash{} & 0.005 & \textbackslash{} & 0.4  & 0.001     & 0.0001 & 0.00005
\\ \hline
\end{tabular}}
\caption{Experiment settings of our benchmark.}
\label{tab:method_settings}
\end{table*}

The training settings for different methods are presented in Table~\ref{tab:method_settings}. For methods with official source code available, we followed their procedures exactly as outlined in the official documentation. However, for methods without official source code, we used a similar approach to the one used in IM-IAD~\cite{xie2023iad}, utilizing some non-official code for reproduction. Among them, we employed a replay-based training approach for PatchCore and UniAD. We set their cache pool size to 100 training samples, a storage capacity that significantly exceeds that of our key-prompt-knowledge structure. 

For our method, the key-prompt-knowledge structure comprised a key of size (15, 196, 1024) float array, a prompt of size (15, 5, 768) float array, and knowledge of size (15, 196, 1024) float array, with an overall size of approximately 23.19MB. Where 15 represents the maximum number of dataset categories, 1024 signifies the dimension of the 768-dimension features extracted by ViT after mapping, and 196 represents the count of embeddings obtained after flattening the features with (14,14,768) size of a (224,224) size image.
5 is the number of Vit encoder layer. We set the size of the knowledge as (15, 196, 768) with the expectation of covering the most essential features of different categories of items as comprehensively as possible. We set the size of the key to be the same as the knowledge based on similar considerations. However, in reality, the key is only used for categories classification, so reducing its size does not affect the effectiveness of query process.

\subsection{Detail experiment results}

\begin{table*}[!t]
\resizebox{\textwidth}{!}{
\begin{tabular}{lllllllllllllllllll}
\hline
Metric                       & CPM & SCL & Bottle & cable & capsule & carpet & grid  & hazelnut & leather & metal\_nut & pill  & screw & tile  & toothbrush & transistor & wood  & zipper & average \\ \hline
\multirow{3}{*}{Image AUROC} &     &     & 0.502  & 0.551 & 0.343   & 0.925  & 0.709 & 0.910    & 0.999   & 0.596      & 0.481 & 0.487 & 0.819 & 0.814      & 0.383      & 0.932 & 0.938  & 0.693   \\
                             & \checkmark   &     & 0.989  & 0.674 & 0.830   & 0.947  & 0.882 & 0.985    & 1.000   & 0.971      & 0.868 & 0.552 & 0.993 & 0.992      & 0.814      & 0.989 & 0.920  & 0.894   \\
                             & \checkmark   & \checkmark   & 1.000  & 0.751 & 0.866   & 0.965  & 0.944 & 0.994    & 1.000   & 0.988      & 0.894 & 0.739 & 0.998 & 1.000      & 0.874      & 0.995 & 0.938  & 0.930   \\ \hline
\multirow{3}{*}{Pixel AUPR}  &     &     & 0.183  & 0.040 & 0.038   & 0.452  & 0.034 & 0.337    & 0.284   & 0.121      & 0.019 & 0.012 & 0.295 & 0.043      & 0.060      & 0.411 & 0.413  & 0.183   \\
                             & \checkmark   &     & 0.752  & 0.168 & 0.327   & 0.594  & 0.172 & 0.496    & 0.337   & 0.727      & 0.626 & 0.143 & 0.522 & 0.291      & 0.337      & 0.555 & 0.340  & 0.426   \\
                             & \checkmark   & \checkmark   & 0.752  & 0.290 & 0.349   & 0.622  & 0.187 & 0.506    & 0.333   & 0.775      & 0.634 & 0.214 & 0.549 & 0.298      & 0.398      & 0.535 & 0.398  & 0.456  \\ \hline
\end{tabular}}
\caption{Ablation study for CPM and SCL on MVTec AD~\cite{bergmann2019mvtec}.}
\label{tab:abliation_module_mvtec}
\end{table*}

\begin{table*}[!t]
\resizebox{\textwidth}{!}{
\begin{tabular}{lllccccccccccccc}
\hline
Metric                       & CPM & SCL & candle & capsules & cashew & chewinggum & fryum & macaroni1 & macaroni2 & pcb1  & pcb2  & pcb3  & pcb4  & pipe\_fryum & average \\
\multirow{3}{*}{Image AUROC} &     &     & 0.461  & 0.497    & 0.629  & 0.714      & 0.522 & 0.517     & 0.462     & 0.553 & 0.506 & 0.527 & 0.635 & 0.989       & 0.584   \\
                             & \checkmark  &     & 0.635  & 0.756    & 0.847  & 0.944      & 0.905 & 0.680     & 0.612     & 0.883 & 0.771 & 0.717 & 0.717 & 0.960       & 0.786   \\
                             & \checkmark  & \checkmark   & 0.778  & 0.877    & 0.960  & 0.958      & 0.945 & 0.823     & 0.667     & 0.905 & 0.871 & 0.813 & 0.901 & 0.988       & 0.874   \\ \hline
\multirow{3}{*}{Pixel AUPR}  &     &     & 0.001  & 0.009    & 0.022  & 0.029      & 0.030 & 0.000     & 0.000     & 0.011 & 0.007 & 0.006 & 0.010 & 0.475       & 0.050   \\
                             & \checkmark   &     & 0.026  & 0.302    & 0.561  & 0.496      & 0.244 & 0.004     & 0.004     & 0.584 & 0.117 & 0.180 & 0.053 & 0.441       & 0.251   \\
                             & \checkmark   & \checkmark   & 0.067  & 0.437    & 0.580  & 0.503      & 0.334 & 0.013     & 0.003     & 0.702 & 0.136 & 0.266 & 0.106 & 0.457       & 0.300  \\ \hline
\end{tabular}}
\caption{Ablation study for CPM and SCL on VisA~\cite{zou2022spot}.}
\label{tab:abliation_module_visa}
\end{table*}

\begin{table*}[!t]
\resizebox{\textwidth}{!}{
\begin{tabular}{llllcccccccccccccccc}
\hline
Metric                       & CPM & SCL & Knowledge & Bottle & cable & capsule & carpet & grid  & hazelnut & leather & metal\_nut & pill  & screw & tile  & toothbrush & transistor & wood  & zipper & average \\ \hline
\multirow{6}{*}{Image AUROC} & \checkmark   &     & 1x             & 0.989  & 0.674 & 0.830   & 0.947  & 0.882 & 0.985    & 1.000   & 0.971      & 0.868 & 0.552 & 0.993 & 0.992      & 0.814      & 0.989 & 0.920  & 0.894   \\
                             & \checkmark   &     & 2x             & 0.998  & 0.693 & 0.874   & 0.965  & 0.921 & 0.979    & 1.000   & 0.983      & 0.875 & 0.746 & 0.995 & 1.000      & 0.872      & 0.988 & 0.930  & 0.921   \\
                             & \checkmark   &     & 4x             & 0.998  & 0.658 & 0.911   & 0.974  & 0.947 & 0.979    & 1.000   & 0.979      & 0.884 & 0.788 & 0.995 & 1.000      & 0.903      & 0.987 & 0.935  & 0.929   \\
                             & \checkmark   & \checkmark   & 1x             & 1.000  & 0.751 & 0.866   & 0.965  & 0.944 & 0.994    & 1.000   & 0.988      & 0.894 & 0.739 & 0.998 & 1.000      & 0.874      & 0.995 & 0.938  & 0.930   \\
                             & \checkmark   & \checkmark   & 2x             & 1.000  & 0.713 & 0.903   & 0.961  & 0.952 & 0.989    & 1.000   & 0.990      & 0.894 & 0.794 & 0.997 & 1.000      & 0.917      & 0.991 & 0.942  & 0.936   \\
                             & \checkmark   & \checkmark   & 4x             & 0.999  & 0.671 & 0.925   & 0.964  & 0.957 & 0.986    & 1.000   & 0.993      & 0.896 & 0.840 & 0.997 & 1.000      & 0.918      & 0.991 & 0.938  & 0.938   \\ \hline
\multirow{6}{*}{Pixel AUPR}  & \checkmark   &     & 1x             & 0.752  & 0.168 & 0.327   & 0.594  & 0.172 & 0.496    & 0.337   & 0.727      & 0.626 & 0.143 & 0.522 & 0.291      & 0.337      & 0.555 & 0.340  & 0.426   \\
                             & \checkmark   &     & 2x             & 0.749  & 0.255 & 0.343   & 0.628  & 0.184 & 0.514    & 0.337   & 0.749      & 0.623 & 0.188 & 0.544 & 0.296      & 0.407      & 0.546 & 0.413  & 0.452   \\
                             & \checkmark   &     & 4x             & 0.754  & 0.176 & 0.346   & 0.634  & 0.188 & 0.512    & 0.336   & 0.790      & 0.619 & 0.210 & 0.550 & 0.298      & 0.445      & 0.548 & 0.398  & 0.453   \\
                             & \checkmark   & \checkmark   & 1x             & 0.752  & 0.290 & 0.349   & 0.622  & 0.187 & 0.506    & 0.333   & 0.775      & 0.634 & 0.214 & 0.549 & 0.298      & 0.398      & 0.535 & 0.398  & 0.456   \\
                             & \checkmark   & \checkmark   & 2x             & 0.751  & 0.270 & 0.347   & 0.624  & 0.189 & 0.515    & 0.334   & 0.794      & 0.635 & 0.234 & 0.546 & 0.299      & 0.400      & 0.555 & 0.424  & 0.461   \\
                             & \checkmark   & \checkmark   & 4x             & 0.752  & 0.250 & 0.347   & 0.639  & 0.189 & 0.521    & 0.334   & 0.802      & 0.627 & 0.236 & 0.536 & 0.298      & 0.460      & 0.571 & 0.426  & 0.466   \\ \hline
\end{tabular}}
\caption{Ablation study for \textit{Knowledge} size and SCL on MVTec AD~\cite{bergmann2019mvtec}.}
\label{tab:abliation_k_mvtec}
\end{table*}

\begin{table*}[!t]
\resizebox{\textwidth}{!}{
\begin{tabular}{llllccccccccccccc}
\hline
Metric                       & CPM & SCL & Knowledge & candle & capsules & cashew & chewinggum & fryum & macaroni1 & macaroni2 & pcb1  & pcb2  & pcb3  & pcb4  & pipe\_fryum & average \\
\multirow{6}{*}{Image AUROC} & \checkmark   &     & 1x             & 0.635  & 0.756    & 0.847  & 0.944      & 0.905 & 0.680     & 0.612     & 0.883 & 0.771 & 0.717 & 0.717 & 0.960       & 0.786   \\
                             & \checkmark   &     & 2x             & 0.705  & 0.760    & 0.895  & 0.961      & 0.907 & 0.784     & 0.633     & 0.820 & 0.827 & 0.677 & 0.863 & 0.981       & 0.818   \\
                             & \checkmark   &     & 4x             & 0.780  & 0.791    & 0.920  & 0.954      & 0.900 & 0.827     & 0.693     & 0.891 & 0.856 & 0.798 & 0.924 & 0.989       & 0.860   \\
                             & \checkmark   & \checkmark   & 1x             & 0.778  & 0.877    & 0.960  & 0.958      & 0.945 & 0.823     & 0.667     & 0.905 & 0.871 & 0.813 & 0.901 & 0.988       & 0.874   \\
                             & \checkmark   & \checkmark   & 2x             & 0.822  & 0.872    & 0.966  & 0.965      & 0.941 & 0.889     & 0.673     & 0.937 & 0.892 & 0.827 & 0.941 & 0.991       & 0.893   \\
                             & \checkmark   & \checkmark   & 4x             & 0.825  & 0.871    & 0.962  & 0.974      & 0.962 & 0.912     & 0.725     & 0.945 & 0.924 & 0.858 & 0.961 & 0.992       & 0.909   \\ \hline
\multirow{6}{*}{Pixel AUPR}  & \checkmark   &     & 1x             & 0.026  & 0.302    & 0.561  & 0.496      & 0.244 & 0.004     & 0.004     & 0.584 & 0.117 & 0.180 & 0.053 & 0.441       & 0.251   \\ 
                             & \checkmark   &     & 2x             & 0.070  & 0.336    & 0.547  & 0.449      & 0.274 & 0.007     & 0.005     & 0.509 & 0.149 & 0.153 & 0.088 & 0.479       & 0.255   \\
                             & \checkmark   &     & 4x             & 0.077  & 0.422    & 0.593  & 0.462      & 0.282 & 0.008     & 0.007     & 0.656 & 0.148 & 0.243 & 0.136 & 0.489       & 0.294   \\
                             & \checkmark   & \checkmark   & 1x             & 0.067  & 0.437    & 0.580  & 0.503      & 0.334 & 0.013     & 0.003     & 0.702 & 0.136 & 0.266 & 0.106 & 0.457       & 0.300   \\
                             & \checkmark   & \checkmark   & 2x             & 0.079  & 0.461    & 0.587  & 0.487      & 0.348 & 0.013     & 0.009     & 0.677 & 0.163 & 0.270 & 0.134 & 0.457       & 0.307   \\
                             & \checkmark   & \checkmark   & 4x             & 0.083  & 0.468    & 0.596  & 0.475      & 0.337 & 0.014     & 0.009     & 0.684 & 0.163 & 0.256 & 0.177 & 0.458       & 0.310  \\ \hline
\end{tabular}}
\caption{Ablation study for \textit{Knowledge} size and SCL on VisA~\cite{zou2022spot}.}
\label{tab:abliation_k_visa}
\end{table*}

\begin{table*}[!b]
\resizebox{\textwidth}{!}{
\begin{tabular}{lccccccccccccccccc}
\hline
Metric                       & ViT Layer & Bottle & cable & capsule & carpet & grid  & hazelnut & leather & metal\_nut & pill  & screw & tile  & toothbrush & transistor & wood  & zipper & average \\ \hline
\multirow{5}{*}{Image AUROC} & 1             & 0.993  & 0.593 & 0.834   & 0.842  & 0.880 & 0.973    & 1.000   & 0.764      & 0.867 & 0.262 & 0.979 & 0.994      & 0.713      & 0.989 & 0.919  & 0.840   \\
                             & 3             & 0.996  & 0.619 & 0.926   & 0.961  & 0.962 & 0.998    & 1.000   & 0.968      & 0.942 & 0.753 & 0.999 & 1.000      & 0.937      & 0.997 & 0.954  & 0.934   \\
                             & 5             & 1.000  & 0.751 & 0.866   & 0.965  & 0.944 & 0.994    & 1.000   & 0.988      & 0.894 & 0.739 & 0.998 & 1.000      & 0.874      & 0.995 & 0.938  & 0.930   \\
                             & 7             & 0.999  & 0.847 & 0.861   & 0.958  & 0.903 & 1.000    & 1.000   & 0.989      & 0.915 & 0.778 & 1.000 & 0.997      & 0.897      & 0.961 & 0.940  & 0.936   \\
                             & 9             & 1.000  & 0.841 & 0.783   & 0.953  & 0.817 & 0.986    & 1.000   & 0.970      & 0.890 & 0.656 & 0.992 & 0.964      & 0.850      & 0.977 & 0.917  & 0.906   \\ \hline
\multirow{5}{*}{Pixel AUPR}  & 1             & 0.698  & 0.230 & 0.245   & 0.507  & 0.128 & 0.492    & 0.430   & 0.555      & 0.610 & 0.006 & 0.554 & 0.365      & 0.202      & 0.495 & 0.466  & 0.399   \\
                             & 3             & 0.759  & 0.026 & 0.339   & 0.540  & 0.195 & 0.520    & 0.405   & 0.788      & 0.669 & 0.217 & 0.527 & 0.333      & 0.405      & 0.521 & 0.522  & 0.451   \\
                             & 5             & 0.752  & 0.290 & 0.349   & 0.622  & 0.187 & 0.506    & 0.333   & 0.775      & 0.634 & 0.214 & 0.549 & 0.298      & 0.398      & 0.535 & 0.398  & 0.456   \\
                             & 7             & 0.734  & 0.371 & 0.342   & 0.640  & 0.160 & 0.530    & 0.284   & 0.695      & 0.624 & 0.196 & 0.498 & 0.319      & 0.426      & 0.517 & 0.326  & 0.444   \\
                             & 9             & 0.718  & 0.300 & 0.284   & 0.601  & 0.139 & 0.489    & 0.266   & 0.735      & 0.659 & 0.136 & 0.516 & 0.343      & 0.426      & 0.430 & 0.251  & 0.420 \\ \hline 
\end{tabular}}
\caption{Ablation study for ViT encoder layer on MVTec AD~\cite{bergmann2019mvtec}.}
\label{tab:abliation_layer_mvtec}
\end{table*}

\begin{table*}[!b]
\resizebox{\textwidth}{!}{
\begin{tabular}{lcccccccccccccc}
\hline
Metric                       & ViT Layer & candle & capsules & cashew & chewinggum & fryum & macaroni1 & macaroni2 & pcb1  & pcb2  & pcb3  & pcb4  & pipe\_fryum & average \\ \hline
\multirow{5}{*}{Image AUROC} & 1             & 0.787  & 0.844    & 0.970  & 0.959      & 0.918 & 0.860     & 0.612     & 0.868 & 0.706 & 0.669 & 0.508 & 0.969       & 0.806   \\
                             & 3             & 0.875  & 0.895    & 0.971  & 0.975      & 0.937 & 0.910     & 0.675     & 0.896 & 0.879 & 0.693 & 0.826 & 0.981       & 0.876   \\
                             & 5             & 0.778  & 0.877    & 0.960  & 0.958      & 0.945 & 0.823     & 0.667     & 0.905 & 0.871 & 0.813 & 0.901 & 0.988       & 0.874   \\
                             & 7             & 0.840  & 0.848    & 0.980  & 0.945      & 0.938 & 0.833     & 0.613     & 0.907 & 0.867 & 0.781 & 0.917 & 0.992       & 0.872   \\
                             & 9             & 0.813  & 0.791    & 0.969  & 0.945      & 0.905 & 0.822     & 0.624     & 0.874 & 0.852 & 0.758 & 0.909 & 0.978       & 0.853   \\ \hline
\multirow{5}{*}{Pixel AUPR}  & 1             & 0.076  & 0.637    & 0.219  & 0.194      & 0.222 & 0.024     & 0.013     & 0.037 & 0.050 & 0.005 & 0.010 & 0.479       & 0.164   \\
                             & 3             & 0.089  & 0.594    & 0.432  & 0.297      & 0.306 & 0.028     & 0.012     & 0.754 & 0.194 & 0.155 & 0.059 & 0.476       & 0.283   \\
                             & 5             & 0.067  & 0.437    & 0.580  & 0.503      & 0.334 & 0.013     & 0.003     & 0.702 & 0.136 & 0.266 & 0.106 & 0.457       & 0.300   \\
                             & 7             & 0.062  & 0.293    & 0.659  & 0.446      & 0.406 & 0.008     & 0.003     & 0.500 & 0.082 & 0.221 & 0.135 & 0.387       & 0.267   \\
                             & 9             & 0.070  & 0.244    & 0.714  & 0.403      & 0.419 & 0.007     & 0.004     & 0.393 & 0.046 & 0.175 & 0.154 & 0.348       & 0.248   \\ \hline
\end{tabular}}
\caption{Ablation study for ViT encoder layer on VisA~\cite{zou2022spot}.}
\label{tab:abliation_layer_visa}
\end{table*}

Due to the page limitation of the main text, we provide detailed metrics of different categories for ablation study in the supplementary material.

The Tables~\ref{tab:abliation_module_mvtec}-~\ref{tab:abliation_layer_visa} provided here are detailed version for the ablation experiments in the main text. From Tables~\ref{tab:abliation_layer_mvtec} and ~\ref{tab:abliation_layer_visa}, it can be observed that features extracted from different ViT encoder layers have their own advantages for anomaly detection in different categories of objects. We believe that combining these features could lead to even better performance.

\end{document}